\documentclass[runningheads]{llncs}
\usepackage[T1]{fontenc}

\usepackage{hyperref}
\usepackage{amsmath}
\usepackage{amssymb}
\usepackage{algorithm}
\usepackage{algorithmicx}
\usepackage{algpseudocode}
\usepackage{booktabs}
\usepackage{multirow}
\usepackage{xcolor}
\usepackage{subcaption}
\usepackage{enumitem}
\usepackage{graphicx}

\begin{document}
\title{Graph Neural Networks at a Fraction}


\author{
Rucha Bhalchandra Joshi\inst{1}\thanks{This research was done while R. Joshi was a student at NISER, Bhubaneswar\inst{1} and HBNI, Mumbai\inst{2}.} \and
Sagar Prakash Barad\inst{2}\inst{3} \and
Nidhi Tiwari\inst{4} \and
Subhankar Mishra\inst{2}\inst{3}
}

\authorrunning{R. Joshi et al.}

\institute{
The Cyprus Institute, Nicosia, Cyprus \and
National Institute of Science Education and Research, Bhubaneswar, India \and
Homi Bhabha National Institute, Mumbai, India \and
Microsoft Ltd., India
\\
\email{r.joshi@cyi.ac.cy, sagar.barad@niser.ac.in, nidhitiwari@microsoft.com, smishra@niser.ac.in}\\
}
\maketitle  

\begin{abstract}
Graph Neural Networks (GNNs) have emerged as powerful tools for learning representations of graph-structured data. In addition to real-valued GNNs, quaternion GNNs also perform well on tasks on graph-structured data. With the aim of reducing the energy footprint, we reduce the model size while maintaining accuracy comparable to that of the original-sized GNNs.
This paper introduces Quaternion Message Passing Neural Networks (QMPNNs), a framework that leverages quaternion space to compute node representations. Our approach offers a generalizable method for incorporating quaternion representations into GNN architectures at one-fourth of the original parameter count.
Furthermore, we present a novel perspective on Graph Lottery Tickets, redefining their applicability within the context of GNNs and QMPNNs. We specifically aim to find the initialization lottery from the subnetwork of the GNNs that can achieve comparable performance to the original GNN upon training. Thereby reducing the trainable model parameters even further.
To validate the effectiveness of our proposed QMPNN framework and LTH for both GNNs and QMPNNs, we evaluate their performance on real-world datasets across three fundamental graph-based tasks: node classification, link prediction, and graph classification. Our code is available at project's \href{https://github.com/SagarPrakashBarad/QuatGLT}{GitHub repository}.

\keywords{Graph Neural Networks  \and Pruning \and Quaternion Graph Neural Networks \and Lottery Ticket Hypothesis.}
\end{abstract}

\section{Introduction}
\label{sec:introduction}
Quaternions, a hypercomplex number system, offer several advantages over traditional real and complex numbers, making them well-suited for various tasks in deep learning. Incorporating quaternions into deep neural networks holds significant promise for enhancing their expressive power and versatility. When used in a quaternion setting, Quaternion deep neural networks perform better than their natural counterparts \cite{parcollet2020survey}. The added advantage of the quaternions is that they reduce the degree of freedom of parameters by one-fourth, which effectively means that only one-fourth of the parameters must be tuned. The inherent multidimensional nature of quaternions, coupled with their capacity to diminish parameter counts significantly, suggests that hyper-complex numbers are more suitable than real numbers for crafting more efficient models in multidimensional spaces with lesser trainable parameter counts.

Most GNNs learn the embeddings in the Euclidean space. 
A recent approach \cite{nguyen2021quaternion} proposes learning graph embeddings in Quaternion space. However, it is the quaternion version of one model GCN \cite{kipf2016semi}, and it cannot be generalized further. The GNN variants exploit different interesting properties of the graphs to learn the representations efficiently. A quaternion framework should be able to maintain these characteristics of the GNN model and embrace them in the quaternion space. With this realization, in this work, we present a generalizable quaternion framework that can adapt to any GNN variant.

Additionally, unlike in the case of the non-GNN deep neural networks \cite{iqbal2023neural,mukhopadhyay2024large,mukhopadhyay2025transformers}, the existing QGNNs do not leverage the quaternion space to reduce the number of parameters. In these non-GNN methods, the input features are usually divided in such a way that overall the four components of the quaternion vector can aggregate them and make a meaningful transformation out of it. The previous work \cite{nguyen2021quaternion}, however, uses each quaternion component to work with all the features. This adds four times the computational overhead, making the model expensive in terms of training and inference time. Additionally, it makes the outputs very sensitive to even a tiny change in the input, as this type of quaternion GNN amplifies the tiny change.  To this end, we restrict our quaternion model from quadrupling the number of parameters, thereby reducing the trainable parameter count to one-fourth of that of real, and this also prohibits the output from being extremely sensitive to the tiny change in the input features. 


\begin{figure*}[ht]
    \centering
    \includegraphics[width=\linewidth]{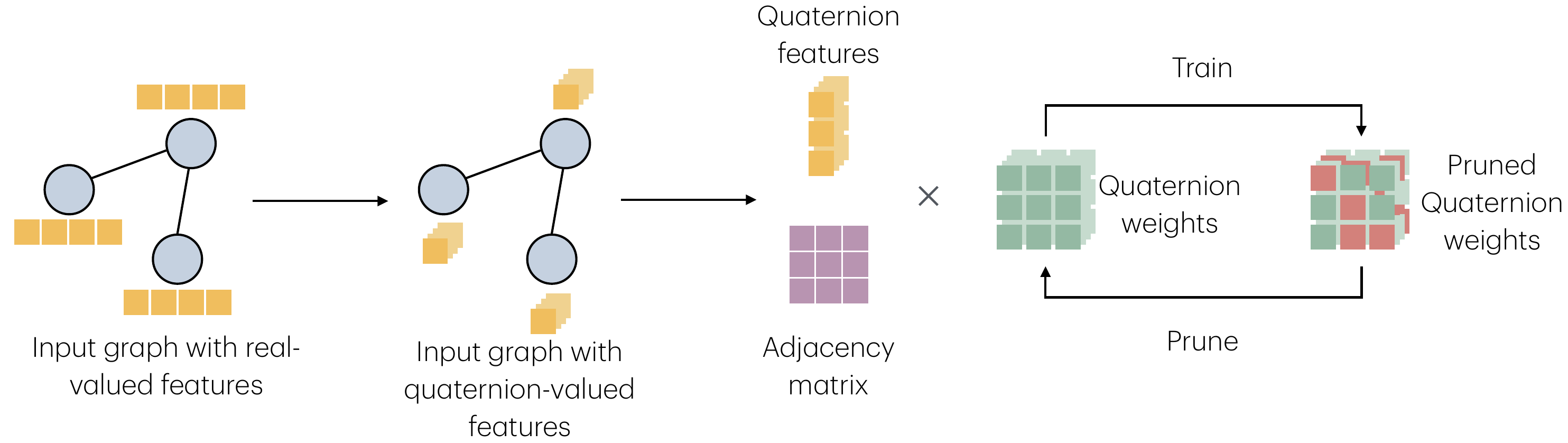}
    \caption{Pruning a quaternion message passing neural network. Graph's features are first transformed into quaternion features. We train QMPNN, which has quaternion weights, for the given task. Furthermore, to find our proposed winning lottery ticket, we prune and train the QMPNN until we get a model with pruned quaternion weights that gives a comparable accuracy. The pruned quaternion network can be trained with only a small fraction of the parameters in the real GNN.
    }
    \label{fig:qmpnn}
\end{figure*}

In this work, our goal is to attain Quaternion Graph Neural Networks that are expressive and have fewer parameters. To do this, we \emph{first} propose a generic framework to get the Quaternion model corresponding to any GNN model with real representations. \emph{Secondly}, we present a framework to sparsify the Quaternion GNNs so that we reduce the number of parameters by finding a lottery ticket.
We generalize the Lottery Ticket Hypothesis \cite{frankle2018lottery} to Quaternion Graph Neural Networks.  For graphs, the existing unified graph lottery ticket, which prunes the input graph as well as the GNN to obtain the winning graph lottery ticket, has some drawbacks. Primarily, it is not feasible to prune the input graph along with the GNNs in graph-level tasks. UGS lacks effectiveness for transfer learning purposes because of its dependency on the input graph. We tackle this by redefining the lottery ticket for graphs. Figure \ref{fig:qmpnn} presents an overview of our approach.


To summarize, this paper makes contributions as follows:
(1) Quaternion Message Passing Neural Networks (QMPNNs): We present a generalizable framework for computing quaternion space representations using any GNN, reducing trainable parameters by 75\%.
(2) We redefine the graph lottery tickets from a different perspective for GNNs and QMPNNs.
(3) To our knowledge, we are the first to empirically demonstrate the existence of graph lottery tickets in QMPNNs.
(4) We provide the performance evaluation of the proposed LTH on both GNNs and QMPNNs on real-world datasets for three graph-based tasks - node classification, link prediction, and graph classification. 



\section{Background}
\label{sec:background}

\subsection{Notations and Definitions}
We use the following notations in this paper.
A graph $\mathcal{G} = \{ \mathcal{V}, \mathcal{E}\}$ has a set of vertices $\mathcal{V}$ and a set of edges $\mathcal{E}$. The feature vector corresponding to a node $v \in \mathcal{V}$ is given as $\mathrm{x}_v$ in $\mathbb{R}^F$. The feature matrix for the graph is given as $\mathbf{X} \in \mathrm{R}^{|\mathcal{V}| \times F}$. The edge $e_{ij} = (v_i, v_j) \in \mathcal{E}$ if and only if there is an edge between two nodes $v_i$ and $v_j$. In the adjacency matrix representation $\mathbf{A} \in \mathbb{R}^{|\mathcal{V}| \times |\mathcal{V}|}$ of the graph, the $\mathbf{A}_{ij} = 1$ if $e_{ij} \in \mathcal{E}$, and $0$ otherwise.

The intermediate representations in layer $l$ corresponding to a node $v$ are given by $h_v^{(l)}$. Similarly, we denote representations in layer $l$ for node $v$ by $h_v^{(l), Q}$, using the additional superscript $Q$ to represent quaternion.

\subsection{Quaternions}
\label{sec:quaternions}
The set of quaternions is denoted by $\mathbb{H}$. A quaternion $q$ is denoted as $q = q_r + q_i \mathrm{i} + q_j \mathrm{j} + q_k \mathrm{k}$, where $q_r, q_i, q_j, q_k \in \mathbb{R}$ and $\mathrm{i}, \mathrm{j}, \mathrm{k}$ are imaginary units. They follow the relation $\mathrm{i}^2 = \mathrm{j}^2 = \mathrm{k}^2 = \mathrm{ijk} = -1$. 

$\mathbf{Addition}$: The quaternion addition of two quaternoons $q$ and $p$ is component-wise: $(q_r + q_i \mathrm{i} + q_j \mathrm{j} + q_k \mathrm{k}) + (p_r + p_i \mathrm{i} + p_j \mathrm{j} + p_k \mathrm{k}) = (q_r + p_r) + (q_i + p_i) \mathrm{i} + (q_j + p_j) \mathrm{j} + (q_k + p_k) \mathrm{k}$. 

$\mathbf{Multiplication}$: Quaternions can be multiplied by each other. The multiplication is associative, i.e., $(pq)r = p(qr)$ for $p, q, r \in \mathbb{H}$. It is also distributive, i.e., $ (p+q)r = pr + qr$. However, the multiplication is not commutative, i.e., $pq \neq qp$. When multiplied with the scalar $\lambda$, it gives $\lambda q = \lambda q_r + \lambda q_i \mathrm{i} + \lambda q_j \mathrm{j} + \lambda q_k \mathrm{k}$. Multiplication of two quaternions $q$ and $p$ is defined by the Hamilton product $q\otimes p$. 
This is given in the matrix form as follows:
\begin{gather}
\label{eq:qmult}
    q \otimes p = 
    \begin{bmatrix}
        1 \\ \mathrm{i} \\ \mathrm{j} \\ \mathrm{k}
    \end{bmatrix}^\top 
    \begin{bmatrix}
        q_r & -q_i & -q_j & -q_k \\
        q_i & q_r & -q_k & q_j \\
        q_j & q_k & q_r & -q_i \\
        q_k & -q_j & q_i & q_r
    \end{bmatrix}
    \begin{bmatrix}
        p_r \\ p_i \\ p_j \\ p_k
    \end{bmatrix}
\end{gather}

$\mathbf{Conjugation}$: $\Bar{q} = q_r - q_i \mathrm{i} - q_j \mathrm{j} - q_k \mathrm{k}$ is the conjugation of the quaternion $q$ given above.

$\mathbf{Norm}$: The norm of quaternion $q$ is given as $\lVert q \rVert = \sqrt{q_r^2 + q_i^2 + q_j^2 + q_k^2}$.

\section{Quaternion Message Passing Neural Networks}
\label{sec:method_qmpnn}

Graph Neural Networks (GNNs) analyze graph-structured data by updating node representations through message passing. At each layer, messages from neighboring nodes are aggregated and used to update node features. The process is formalized as:
\begin{align}
    m_{uv}     &= \textsf{MESSAGE} (h^{(l)}_u, h^{(l)}_v, e_{uv}) \label{eq:message} \\
    h'_u       &= \textsf{AGGREGATE} (m_{uv}), \forall v \in \mathcal{N}(u) \label{eq:aggregate} \\
    h^{(l+1)} &= \textsf{UPDATE} (h^{(l)}_u, h'_u) \label{eq:update}
\end{align}
Here, \textsf{MESSAGE} computes messages based on node features $h^{(l)}_u, h^{(l)}_v$ and edge features $e_{uv}$; \textsf{AGGREGATE} collects messages from neighbors $\mathcal{N}(u)$; and \textsf{UPDATE} refines the node representation. These functions, often learnable and differentiable, enable GNNs to be trained via backpropagation.

Quaternion Graph Neural Networks (QGNNs) refer to neural networks that utilize quaternion weights to perform computations within the quaternion vector space. Notably, QGNN, as developed by \cite{nguyen2021quaternion} is a quaternion variant of the Graph Convolutional Network (GCN) layer only. 
This is the first limitation of QGNN, as it cannot be generalized to other GNN variants. The second limitation of QGNN is by construction, number of trainable parameters remain the same as that of real-valued GCN.
While leveraging the quaternion space to generate expressive representations for nodes, we also seek to take advantage of the quaternions by reducing the number of trainable parameters. To this end, we propose a generalized framework to give the quaternion equivalent of any graph neural network method.

In the quaternion variant of any GNNs, in the equations \ref{eq:message}, \ref{eq:aggregate}, \ref{eq:update}, we ensure that the \textsf{MESSAGE}, \textsf{AGGREGATE} and \textsf{UPDATE} functions follow the quaternion operations as given in section \ref{sec:quaternions}. We describe this in more detail below:

To begin, we have the features represented using the quaternions. \(F\) is the number of features corresponding to every node. To use these features with quaternions, we make sure that \(F\) is divisible by four so that we can associate \(F/4\) real-valued features with a real and three imaginary components of quaternion. Consequently, the feature vector \(h_v^{(l), Q}\) of node \(v\) in quaternion form is represented as \(h_v^{(l), Q} = h^{(l)}_r + h^{(l)}_i \mathrm{i} + h^{(l)}_j \mathrm{j} + h^{(l)}_k \mathrm{k}\).

In every layer \( l\) of GNN, the learnable functions have associated parameters, which we represent using \( W^{(l), Q}\). This weight matrix \(W^{(l), Q}\) has consists of four real-valued components \( W^{(l)}_r,\) \(W^{(l)}_i,\) \(W^{(l)}_j,\) \(W^{(l)}_k \) of the quaternion weight matrix. Specifically, \( W^{(l), Q} = W^{(l)}_r + W^{(l)}_i \mathrm{i} + W^{(l)}_j \mathrm{j} + W^{(l)}_k \mathrm{k} \).

The weights are multiplied with feature maps using the quaternion multiplication rule given in equation \ref{eq:qmult}. Following this, it is done as follows:
\begin{equation}
\label{eq:feat_mult}
    W^{(l+1), Q} = W^{(l), Q} \otimes h_v^{(l), Q}
\end{equation}

\begin{gather}
\label{eq:quat_mat}
    \begin{bmatrix}
        \mathcal{R}(W^{(l), Q}) \\
        \mathcal{I}(W^{(l), Q}) \\
        \mathcal{J}(W^{(l), Q}) \\
        \mathcal{K}(W^{(l), Q})
    \end{bmatrix} =
    \begin{bmatrix}
        W^{(l)}_r & -W^{(l)}_i & -W^{(l)}_j & -W^{(l)}_k \\
        W^{(l)}_i & W^{(l)}_r & -W^{(l)}_k & W^{(l)}_j \\
        W^{(l)}_j & W^{(l)}_k & W^{(l)}_r & -W^{(l)}_i \\
        W^{(l)}_k & -W^{(l)}_j & W^{(l)}_i & W^{(l)}_r \\
    \end{bmatrix} \times
    \begin{bmatrix}
        h^{(l)}_r \\ 
        h^{(l)}_i \\
        h^{(l)}_j \\
        h^{(l)}_k \\
    \end{bmatrix}
\end{gather}
where \(\mathcal{R, I, J, K}\) denote the four imaginary components of the product of weights with the node representations. 

Due to quaternion weights being multiplied with quaternion features, the degrees of freedom are reduced by one-fourth. The product in equation \ref{eq:feat_mult} combines components \( (r, i, j, k) \), capturing complex interdependencies. Each quaternion weight, as shown in equation \ref{eq:quat_mat}, encodes spatial correlations across features, resulting in more expressive networks compared to real-valued counterparts~\cite{trabelsi2018deep}. Additionally, the trainable parameters are reduced to one-fourth of those in an equivalent GNN with real weights, aiding in model size reduction.


\subsection{Differentiability}
For the QMPNNs to learn, the quaternion versions of learnable functions from \textsf{MESSAGE}, \textsf{AGGREGATE}, and \textsf{UPDATE} should be differentiable. For our framework, the differentiability follows from the generalized complex chain rule for a real-valued loss function, which is provided in much detail in Deep Complex Networks \cite{trabelsi2018deep} and Deep Quaternion Networks \cite{gaudet2018deep}. 

\subsection{Computational Complexity}
Four feature values are clubbed together to form a single quaternion in QMPNNs, hence \(F/4\) quaternions are necessary to transform the feature vector. The transformed feature space is of dimension \(F'\). While using quaternion weights, these are considered as the combination of \(F'/4\) quaternions. The number of parameters of the quaternion weight matrix, hence, is \(F/4 \times F'/4\), with each of them having four quaternion components. Since
the components \( W_r^{(l)}, W_i^{(l)}, W_j^{(l)},\) and \( W_k^{(l)}\) of the quaternion weight \( W^{(l), Q}\) are shared during the Hamilton product, the degree of freedom is reduced by one-fourth of that of the real weight matrix. 
For quaternion matrix multiplications has same time complexity as that of the real ones, we reduce the number of trainable parameters without compromising on the time complexity due to use of quaternion parameters.
\\

\section{Lottery Ticket Hypothesis on QMPNNs}
\label{sec:method_lth}

\emph{The lottery ticket hypothesis} (LTH) \cite{frankle2018lottery} states that \emph{a randomly initialized, dense neural network contains a sub-network that is initialized such that—when trained in isolation—it can match the test accuracy of the original network after training for at most the same number of iterations.}

Different from the lottery tickets defined previously \cite{chen2021unified,tsitsulin2023the,wang2023searching,zhang2024graph} - that consider a subgraph of a graph with or without the subnetwork of the GNN as a ticket, we define the LTH for Graph Neural Networks and Quaternion GNNs. 
In a graph neural network, \( f(x; W) \), with initial parameters \( W^{(0)} \), using Stochastic Gradient Descent (SGD), we train on the training set, minimizes to a minimum validation loss \( l \) in \( j \) number of iterations achieving accuracy \( a\). The lottery ticket hypothesis is that there exists \(m\) when optimizing with SGD on the same training set, \( f\) attains the test accuracy \(a' \geq a\) with a validation loss \(l' \leq l\) in \(j' \leq j\) iterations, such that \( \lVert m \rVert \ll \lVert W \rVert\), thereby reducing the number of parameters. The sub-network with parameters \(m \odot W^Q \) is the winning lottery ticket, with the mask \(m\). Algorithm \ref{alg:gnn_sparsification} sparsifies the Graph Neural Network which is required to find the winning lottery ticket. The iterative algorithm that finds it is described in algorithm \ref{alg:iterative_glt}.

\begin{algorithm}
    \caption{GNN Sparsification}
    \label{alg:gnn_sparsification}
    \begin{algorithmic}[1]
        \Require Input graph $\mathcal{G} = (\mathcal{V}, \mathcal{E}, \mathbf{X})$, GNN's initialization $W^{(0)}$, initial mask $m^0 = 1 \in \mathbb{R}^{\lVert W \rVert}$, step size $\eta, \lambda$
        \Ensure Sparsified weight mask $m$
        \For{$i \gets 0$ to $N-1$}
            \State Forward $f(\cdot, m^i \odot W)$ with $\mathcal{G} = (\mathcal{V}, \mathcal{E}, \mathbf{X})$ to compute the loss $\mathcal{L}_{\mathrm{task}}$
            \State Back-propagate to update $W^{(i+1)} \gets W^{(i)} - \eta \nabla_W \mathcal{L}_{\mathrm{task}}$
            \State Update $m^{i+1} \gets m^{i} - \lambda \nabla_{m^i} \mathcal{L}_{\mathrm{task}}$
        \EndFor
        \State Set $p=20\%$ of the lowest magnitude values in $m^N$ to $0$ and others to 1, get mask $m$
    \end{algorithmic}
\end{algorithm}

We primarily differ from the definition of GLT in the previous works, as we observe that this is not suitable for graph-level tasks \cite{sun2023all,nguyen2021quaternion}. Popular methods such as UGS \cite{chen2021unified}, AdaGLT \cite{zhang2024graph} consider only the node-classification and the link prediction task. In graph-level tasks, sparsifying the input graph does not have any significance.
For graph-level tasks, the structure and the features in the entire graph are considered, unlike in the case of node- or edge-level tasks where only the nodes in the local neighborhood and their features are considered by a graph neural network.
Also, in the case of transfer learning, where the pruned sub-network is to be further finetuned for a different but related task, the dependency of the winning ticket on the input graph restricts us from fine-tuning it.

\subsection{Quaternion Graph Lottery Tickets}
Similar to finding a winning lottery ticket in GNNs, given a QMPNN \(f(\cdot, W^{Q})\) and a graph \(\mathcal{G} = (\mathcal{V}, \mathcal{E})\), the subnetwork, i.e., winning lottery ticket, of the QMPNN is defined as \(f(\cdot, m \odot W^Q)\), where \(m\) is the binary mask on parameters. To find the winning lottery ticket in a QMPNN, we use the algorithms \ref{alg:gnn_sparsification} and \ref{alg:iterative_glt} with the quaternion-values model weights and input features. 


\begin{algorithm}
    \caption{Iterative algorithm to find Graph Lottery Ticket}
    \label{alg:iterative_glt}
    \begin{algorithmic}[1]
        \Require Input graph $\mathcal{G} = (\mathcal{V}, \mathcal{E}, \mathbf{X})$, GNN's initialization $W^{(0)}$, initial mask $m^0 = 1 \in \mathbb{R}^{\lVert W \rVert}$, sparsity level $s$
        \Ensure Lottery Ticket $f(({\mathcal{V}, \mathcal{E}, \mathbf{X}}), m \odot W^0)$
        \While{$1 - \frac{\lVert m \rVert}{\lVert W \rVert}$}
            \State \label{step:gnn_sparsification} Sparsify GNN $f(\cdot, m \odot W^{0})$ using algorithm \ref{alg:gnn_sparsification}
            \State Update $m$ according to step \ref{step:gnn_sparsification}
            \State Reset GNN's weight to $W^0$
        \EndWhile
    \end{algorithmic}
\end{algorithm}

\section{Experiments}
\label{sec:expt}
The effectiveness of QMPNNs for different GNN variants is validated with extensive experimentation on different real-world standard datasets. We also verify the existence of the GLTs as per our definition. We evaluated their performance on three tasks, node classification, link prediction, and graph classification.


\subsection{Datasets}
\label{sec:datasets}
Tables \ref{tab:node-link-data} and \ref{tab:graph-data} summarize the datasets and their statistics. For quaternion models, dataset features and classes were adjusted to be divisible by 4 by padding features with the average value and adding dummy classes. For instance, Cora's feature size was padded from 1,433 to 1,436, and the number of classes was increased from 7 to 8.

\begin{table}[h]
\centering
\caption{Node classification and link prediction datasets statistics}
\resizebox{0.9\textwidth}{!}{%
\begin{tabular}{lrrrcl}
\toprule
\textbf{Dataset} & \textit{\#Nodes} & \textit{\#Edges} & \textit{\#Features} & \textit{\#Classes} & \textit{Metric} \\ \midrule
Cora        & 2,708    & 5,429    & 1,433    & 7  & Accuracy, ROC-AUC \\ 
Citeseer    & 3,327    & 4,732    & 3,703    & 6  & Accuracy, ROC-AUC\\ 
PubMed      & 19,717   & 44,338   & 500      & 3  & Accuracy, ROC-AUC\\ 
ogbn-arxiv  & 169,343  & 1,166,243 & 128     & 40 & Accuracy, ROC-AUC\\ 
ogbl-collab & 235,868  & 2,358,104 & 128     & 0  & Hits@50  \\ 
\bottomrule
\end{tabular}
}
\label{tab:node-link-data}
\end{table}


\begin{table}[h]
\centering
\caption{Graph classification datasets statistics}
\begin{tabular}{lrrcc}
\toprule
\textbf{Dataset} & \textit{\#Graphs} & \textit{Avg. \#Edges} & \textit{\#Classes} & \textit{Metric} \\ \midrule
MUTAG      & 188    & 744     & 2  & Accuracy \\ 
ENZYMES    & 600    & 1,686   & 6  & Accuracy \\ 
PROTEINS   & 1,113  & 3,666   & 2  & Accuracy \\ 
ogbg-molhiv  & 41,127 & 40      & 2  & Accuracy \\ \bottomrule
\end{tabular}
\label{tab:graph-data}
\end{table}

\subsection{Experimental Setup}

\begin{figure*}[!ht]
    \centering
    \includegraphics[width=\textwidth]{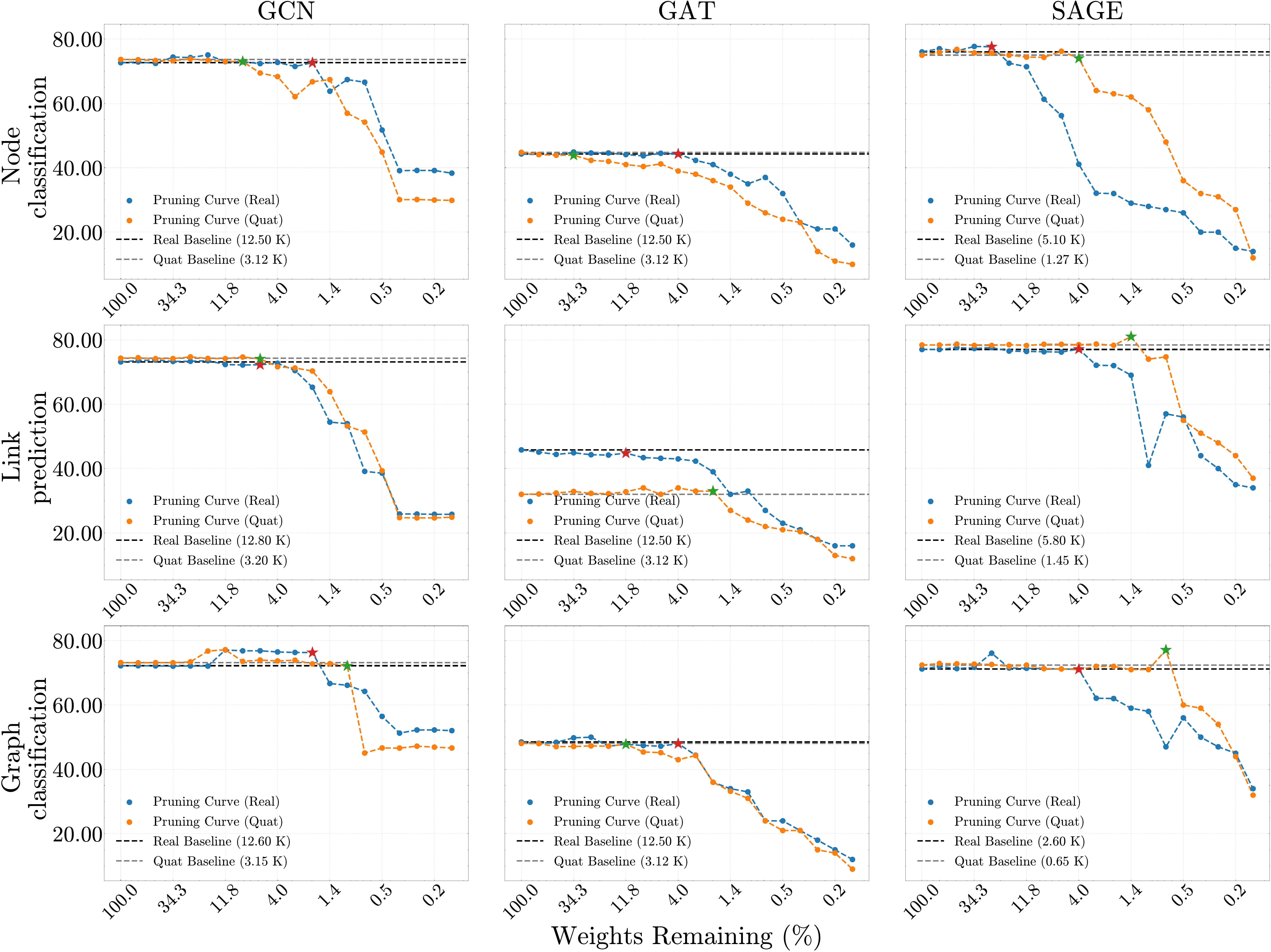}
    \caption{\textbf{Performance After Pruning}: The plots show GCN, GAT, and GraphSAGE performance on OGBN-ARXIV (node classification), OGBL-COLLAB (link prediction), and OGBG-MOHLIV (graph classification) at a pruning weight fraction of 0.3. \textbf{GLTs} are marked by red (\textcolor{red}{$\star$}) and green (\textcolor{green}{$\star$}) stars, indicating comparable performance despite sparsity, while dashed lines represent pre-pruning baselines. The plot sections correspond to node classification (top), link prediction (middle), and graph classification (bottom).
    }
    \label{fig:pruning}
\end{figure*}

\begin{itemize}[topsep=0pt, itemsep=2pt, parsep=0pt]
    \item \textbf{GNN Architecture:}  
    Two-layer GCN, GAT, and GraphSAGE models with 128 hidden units were used for node classification and link prediction on Cora, Citeseer, PubMed, ogbn-arxiv, and ogbn-collab. For graph classification on MUTAG, PROTEINS, ENZYMES, and ogbg-molhiv, three-layer models with 128 hidden units per layer were employed.
    
    \item \textbf{Training and Hyperparameters:}  
    Node classification datasets followed an 80-10-10 split, while link prediction and graph classification used an 85-5-10 split. Models were trained with a learning rate of 0.01, weight decay of \(5 \times 10^{-4}\), dropout rate of 0.6, and the Adam optimizer, for up to 1000 epochs with early stopping after 200 epochs of no improvement. A prune fraction of 0.3 was applied.
    
    \item \textbf{Evaluation and Infrastructure:}  
    Evaluation metrics are provided in Tables \ref{tab:node-link-data} and \ref{tab:graph-data}. Experiments were conducted on NVIDIA RTX 3090 GPUs with 24GB VRAM.
\end{itemize}

\subsection{Training and Inference Details}
Our evaluation metrics, detailed in Tables \ref{tab:node-link-data} and \ref{tab:graph-data} and following \cite{kipf2016semi}, include both accuracy and ROC-AUC. Accuracy measures the proportion of correctly classified instances among the total instances, providing a straightforward metric for model performance. The ROC-AUC (Receiver Operating Characteristic - Area Under the Curve) score, representing the degree of separability, is particularly valuable for assessing performance on imbalanced datasets, such as in link prediction tasks with a substantial class imbalance between positive and negative edges. Summarizing the results across these diverse tasks, tables \ref{tab:node_classification}
, \ref{tab:link_prediction}, and \ref{tab:graph_classification} present the inference outcomes for all models, facilitating comprehensive analysis.

\subsection{Results}

GCN, GAT, and GraphSAGE results on Cora, Citesser, PubMed, and ogbn-arxiv for node classification are collected in table \ref{tab:node_classification} and \ref{tab:link_prediction} shows results for link prediction on Cora, Citesser, PubMed, and ogbl-collab. The results pertaining to graph classification tasks on MUTAG, PROTEINS, ENZYMES, and ogbg-molhiv datasets are presented in table \ref{tab:graph_classification}. These tables display the accuracy and parameter count (Params) in millions for each model and dataset, with better-performing models highlighted in bold for easy identification. Pruning results for the same models on ogbn-arxiv for node classification are shown in the top section of figure \ref{fig:pruning}. The middle section of figure \ref{fig:pruning} presents results for link prediction on the ogbl-collab dataset. The bottom section of figure \ref{fig:pruning} illustrates the results for graph classification tasks on the ogbg-molhiv dataset. An extensive list of results is available at project's \href{https://github.com/SagarPrakashBarad/QuatGLT}{GitHub} repository.

\begin{table*}[!htbp]
\centering
\caption{\textbf{Node classification} results for GCN, GAT, and GraphSAGE models on semi-supervised graph datasets, including Cora, Citeseer, PubMed, and OBGN-ARXIV.  Accuracy values are reported with their standard deviations, along with the corresponding GFLOPs (denoted as GF) and parameter counts (in K units, denoted as GF).}
\resizebox{\linewidth}{!}{%
\begin{tabular}{ccccccccccccc}
\toprule
\textbf{Model}       & \multicolumn{3}{c}{\textbf{Cora}} & \multicolumn{3}{c}{\textbf{Citeseer}} & \multicolumn{3}{c}{\textbf{PubMed}} & \multicolumn{3}{c}{\textbf{OGBN-Arxiv}} \\  \cmidrule(l){2-13} 
                     & Accuracy     & Par (K)   & GF & Accuracy     & Par (K)    & GF & Accuracy    & Par (K)     & GF & Accuracy      & Par (K)   & GF \\ \midrule
GCN                  & \textbf{85.2 ± 1.3} & 140.5  & 0.33 & 73.5 ± 1.8  & 125.7   & 0.29 & 79.1 ± 1.2  & 131.9   & 0.31 & \textbf{72.8 ± 1.5} & 145.1   & 0.42 \\
QGCN                 & 84.1 ± 1.5         & 42.1   & 0.11 & 72.6 ± 1.3  & 39.2    & 0.09 & 78.4 ± 1.4  & 41.7    & 0.10 & 72.3 ± 1.6         & 43.5    & 0.12 \\
GAT                  & \textbf{87.3 ± 1.0} & 143.2  & 0.37 & 75.4 ± 1.9  & 130.8   & 0.34 & 80.7 ± 1.6  & 136.3   & 0.36 & 72.5 ± 1.2         & 149.5   & 0.45 \\
QGAT                 & 86.4 ± 1.7         & 43.7   & 0.14 & 74.8 ± 1.1  & 41.6    & 0.12 & 79.8 ± 1.5  & 42.9    & 0.13 & \textbf{72.9 ± 1.8} & 44.5    & 0.15 \\
SAGE                 & 85.5 ± 1.2         & 72.6   & 0.23 & \textbf{74.9 ± 1.4} & 66.1    & 0.19 & 80.2 ± 1.9  & 68.7    & 0.20 & \textbf{73.0 ± 1.0} & 75.3    & 0.26 \\
QSAGE                & 84.9 ± 0.8         & 22.1   & 0.08 & 74.3 ± 1.5  & 20.5    & 0.07 & \textbf{80.5 ± 1.3} & 21.3    & 0.08 & 72.8 ± 1.4         & 23.6    & 0.10 \\ \bottomrule
\end{tabular}%
}
\label{tab:node_classification}
\end{table*}

The results indicate that quaternion models (QGCN, QGAT, and QSAGE) perform equally well or better than their real counterparts (GCN, GAT, and GraphSAGE) under the same hyperparameters and training parameters. For instance, QGCN shows slightly improved accuracy on the PubMed dataset, and QGAT performs nearly as well as GAT on Cora and Citeseer for node classification tasks. Moreover, quaternion models consistently outperform their real counterparts in other tasks, including link prediction and graph classification, demonstrating superior performance in almost all cases. Furthermore, it is consistently shown that even when the real models perform better than QMPNNs, they are mostly ahead by only marginal values.

\begin{table*}[!htbp]
\centering
\caption{\textbf{Link prediction} on various datasets including CORA, CITESEER, PUBMED, and OBGL-COLLAB. ROC-AUC and Hits@50 values are reported with their standard deviations, along with the corresponding GFLOPs (denoted as GF) and parameter counts (in K units, denoted as GF).}
\resizebox{\textwidth}{!}{%
\begin{tabular}{ccccccccccccccc}
\toprule
\multirow{2}{*}{Model Name} & \multicolumn{3}{c}{\textbf{CORA}} & \multicolumn{3}{c}{\textbf{CITESEER}} & \multicolumn{3}{c}{\textbf{PUBMED}} & \multicolumn{3}{c}{\textbf{OGBL-COLLAB}} \\  \cmidrule(l){2-13} 
                             & ROC-AUC   & Par (K)  & GF & ROC-AUC     & Par (K)   & GF & ROC-AUC    & Par (K)  & GF & Hits@50        & Par (K)  & GF \\ \midrule
GCN                          & 79.27 ± 0.47      & 96.0     & 0.23  & \textbf{78.53 ± 0.45}        & 241.4      & 0.58 & \textbf{87.60 ± 0.043}       & 36.3    & 0.09 & 44.02 ± 1.57       & 12.5       & 0.03 \\
QGCN                         & \textbf{83.55 ± 0.03}      & 27.2    & 0.07 & 78.36 ± 4.13        & 63.6       & 0.15 & 87.66 ± 0.32       & 12.3     & 0.03 & \textbf{45.80 ± 1.35}       & 6.4       & 0.02 \\
GAT                          & 93.89 ± 0.06      & 96.3    & 0.24 & 85.56 ± 0.38        & 241.6      & 0.60 & \textbf{89.32 ± 0.38}       & 36.6     & 0.09 & \textbf{45.51 ± 1.61}       & 12.8       & 0.03 \\
QGAT                         & \textbf{94.99 ± 0.19}      & 27.5    & 0.07 & \textbf{86.59 ± 0.10}       & 63.8       & 0.16 & 84.04 ± 0.54       & 12.6     & 0.03 & 32.03 ± 1.92       & 65.6       & 0.16 \\
SAGE                         & 89.00 ± 0.19       & 48.1    & 0.12 & 81.65 ± 0.13        & 120.5      & 0.30 & \textbf{83.36 ± 0.0189}       & 16.1     & 0.04 & \textbf{48.50 ± 0.88}       & 6.4       & 0.02 \\ 
QSAGE                        & \textbf{89.35 ± 0.26}      & 13.6  & 0.03 & \textbf{81.93 ± 0.04}        & 31.8    & 0.08 & 79.05 ± 0.32       & 6.1  & 0.02 & 48.07 ± 0.56       & 3.2       & 0.01 \\ \bottomrule
\end{tabular}%
}
\label{tab:link_prediction}
\end{table*}

The results presented in the tables comparing real and quaternion models, and are trained and evaluated using five different seeds. For each data point, we calculated the mean and standard deviation based on these evaluations. The better-performing models are determined by performing paired t-tests. We chose the model with the better performance when the difference between $p$-values of the models was found to be statistically significant. 

\begin{table*}[!htbp]
\centering
\caption{\textbf{Graph Classification} on various datasets including MUTAG, PROTEINS, ENZYMES, and OGBG-MOLHIV. Accuracy values are reported with their standard deviations, along with the corresponding GFLOPs (denoted as GF) and parameter counts (in K units, denoted as GF).}
\resizebox{\textwidth}{!}{%
\begin{tabular}{ccccccccccccccc}
\toprule
\multirow{2}{*}{Model Name} & \multicolumn{3}{c}{\textbf{MUTAG }} & \multicolumn{3}{c}{\textbf{PROTEINS }} & \multicolumn{3}{c}{\textbf{ENZYMES }} & \multicolumn{3}{c}{\textbf{OGBG-MOLHIV }} \\  \cmidrule(l){2-13} 
                             & Accuracy   & Par (K)   & GF & Accuracy     & Par (K)    & GF & Accuracy    & Par (K)  & GF & Accuracy        & Par (K)  & GF \\ \midrule
GCN                          & 83.54 ± 0.40      & 4.8     & 0.01  & \textbf{73.57 ± 0.45}        & 9.2      & 0.02 & \textbf{72.96 ± 0.15}       & 5.6    & 0.01 & \textbf{76.24 ± 0.89}       & 5.1       & 0.01 \\
QGCN                         & \textbf{88.61 ± 0.66}      & 4.4    & 0.01 & 71.55 ± 0.42        & 5.5       & 0.01 & 71.92 ± 1.16       & 4.6     & 0.01 & 75.85 ± 0.93       & 4.5       & 0.01 \\
GAT                          & 85.26 ± 0.02      & 5.4    & 0.01 & 72.68 ± 0.93        & 9.4      & 0.02 & \textbf{74.61 ± 0.05}       & 5.8     & 0.01 & 76.24 ± 0.32       & 5.8       & 0.01 \\
QGAT                         & \textbf{87.25 ± 0.18}      & 4.7    & 0.01 & \textbf{73.22 ± 0.02}        & 5.8       & 0.01 & 73.35 ± 0.02       & 4.9     & 0.01 & \textbf{78.16 ± 0.78}       & 4.8       & 0.01 \\
SAGE                         & 82.56 ± 2.67       & 2.5    & 0.01 & \textbf{72.62 ± 0.22}        & 4.7      & 0.01 & 79.65 ± 3.06       & 2.8     & 0.01 & 71.77 ± 0.97       & 2.6       & 0.01 \\ 
QSAGE                        & \textbf{84.26 ± 0.22}      & 2.2  & 0.01 & 73.08 ± 1.36        & 2.7    & 0.01 & \textbf{78.59 ± 0.73}       & 2.3  & 0.01 & \textbf{72.98 ± 0.12}       & 2.2       & 0.01 \\ \bottomrule
\end{tabular}%
}
\label{tab:graph_classification}
\end{table*}

It is worth noting that the performance of real-valued GNN models, on datasets with and without the dataset augmentation mentioned in the section \ref{sec:datasets} is the same. Hence, it serves as a fair baseline to compare the quaternion-valued GNN performance with the real-valued GNN performance on the augmented dataset. We provide the code in the repository linked in the paper.

\subsection{Key Findings}
We summarize the effectiveness of magnitude pruning and quaternion-based models in GNNs:

\begin{enumerate}[topsep=0pt, itemsep=2pt, parsep=0pt]
    \item \textbf{QMPNNs operate at 1/4th parameters of real-valued models:}  
    QMPNNs achieve comparable or better performance than real-valued models with just 1/4th of the parameters, showcasing their efficiency and expressiveness, as shown in Tables \ref{tab:node_classification}, \ref{tab:link_prediction}, and \ref{tab:graph_classification}.
    
    \item \textbf{GLTs exist at 1/5th or smaller of original size:}  
    Magnitude pruning identifies lottery tickets at 1--20\% of the original model size without performance loss across QGCN, QGAT, and QGraphSAGE tasks, as observed in Tables \ref{tab:node_classification} and \ref{tab:link_prediction}.
    
    \item \textbf{GAT and GraphSAGE sparsify well; Cora is pruning-sensitive:}  
    GATs and GraphSAGE produce sparse GLTs due to attention and sampling techniques, while Cora exhibits significant sensitivity to pruning, with performance dropping at half model size, as detailed in the comprehensive experiments available on the \href{https://anonymous.4open.science/r/lth-qmpnn-B143/}{project's GitHub repository}.
    
    \item \textbf{Less sparsity in graph classification GLTs:}  
    Graph classification tasks require less sparse GLTs to capture complex global features, with smaller parameter counts due to the datasets' limited graph size, as shown in Table \ref{tab:graph_classification} and Figure \ref{fig:pruning}.
    
    \item \textbf{Large graphs maintain performance:}  
    QMPNNs scale well to large graphs, as seen with ogbn-arxiv, ogbn-collab, and ogbn-molhiv datasets in Tables \ref{tab:node_classification}, \ref{tab:link_prediction}, and \ref{tab:graph_classification}, delivering comparable or better performance relative to trainable parameters.
\end{enumerate}

\section{Conclusion and Future Work}
\label{sec:conclusion}

This paper introduces Quaternion Message Passing Neural Networks (QMPNNs) as a versatile framework for graph representation learning, leveraging quaternion representations to capture intricate relationships in graph-structured data. The framework generalizes easily to existing GNN architectures with minimal adjustments, enhancing flexibility and performance across tasks like node classification, link prediction, and graph classification. By redefining graph lottery tickets, we identified key subnetworks that enable efficient training and inference, demonstrating the scalability and effectiveness of QMPNNs on real-world datasets. Future work could explore dynamic graphs, hybrid modalities, and improving interpretability to further expand QMPNN capabilities.

\bibliographystyle{splncs04}
\bibliography{sample-base}

\appendix

\end{document}